\documentclass[10pt,twocolumn,letterpaper]{article}

\usepackage{cvpr}              % To produce the CAMERA-READY version
\usepackage{graphicx}
\usepackage{amsmath}
\usepackage{amssymb}
\usepackage{booktabs}
\usepackage{multirow}
\usepackage{graphicx} 
\usepackage[labelfont=bf]{caption}

\newcommand\blfootnote[1]{%
  \begingroup
  \renewcommand\thefootnote{}\footnote{#1}%
  \addtocounter{footnote}{-1}%
  \endgroup
}

\usepackage[pagebackref,breaklinks,colorlinks]{hyperref}

\usepackage[capitalize]{cleveref}
\crefname{section}{Sec.}{Secs.}
\Crefname{section}{Section}{Sections}
\Crefname{table}{Table}{Tables}
\crefname{table}{Tab.}{Tabs.}

\def\cvprPaperID{*****} % *** Enter the CVPR Paper ID here
\def\confName{CVPR}
\def\confYear{2022}

\begin{document}

% \title{\LaTeX\ Author Guidelines for \confName~Proceedings}
% \title{Comparison of Humans and Machines on Extreme Image Transforms}
\title{Extreme Image Transformations Facilitate \\Robust Latent Object Representations}

\author{Girik Malik $^{\dagger}$\\
Northeastern University\\
{\tt\small malik.gi@northeastern.edu}
\and
Dakarai Crowder \\
University of Illinois\\Urbana-Champaign\\
{\tt\small dcrowd3@illinois.edu}
\and
Ennio Mingolla \\ 
Northeastern University \\
{\tt\small e.mingolla@northeastern.edu}
}
\maketitle
\blfootnote{$\dagger$ Corresponding author \\ GM is also affiliated with Labrynthe Pvt. Ltd., New Delhi, India}

%%%%%%%%% ABSTRACT
\begin{abstract}
Adversarial attacks can affect the object recognition capabilities of machines in wild. These can often result from spurious correlations between input and class labels, and are prone to memorization in large networks. While networks are expected to do automated feature selection, it is not effective at the scale of the object. Humans, however, are able to select the minimum set of features required to form a robust representation of an object. In this work, we show that finetuning any pretrained off-the-shelf network with Extreme Image Transformations (EIT) not only helps in learning a robust latent representation, it also improves the performance of these networks against common adversarial attacks of various intensities. Our EIT trained networks show strong activations in the object regions even when tested with more intense noise, showing promising generalizations across different kinds of adversarial attacks.
\end{abstract}

%%%%%%%%% BODY TEXT

\section{Introduction}
\label{sec:intro}
% basic motivation
% talk about the meaning of the object for humans and machines
% link with Ullman's PNAS paper, learning with noise and the use of additional blocks to simulate V1
% write a subsec/para on what are EITs.
% Can also be moved to related works
Humans are able to recognize objects with functional accuracy and confidence in low visibility settings. The performance also remains consistent in case of change in environment, lighting or weather conditions. Humans might find it hard to tell the exact specie of a tree seen through a dust-storm but it is not misclassified as an animal. This is not just based on all the trees they have previously seen, but also because of their ability to extract the salient feature of a tree. Human visual system is efficient at recognizing the non-redundant backbone features of an object instead of equating the object class to unrelated external conditions.

Multiple studies have probed the lower bound on the number of features required to describe the action/object in the image. Ullman et al. find MIRC which is the smallest possible image size required to identify the object in the image ---- requiring the effective use of all the information in the image~\cite{ullman2016atoms}. The redundant information in visual input can be effective against occlusion, but does not contribute much for solving natural visual tasks in the wild. 

Our work is motivated by the fact that such minimal features can be used to learn the right latent representation of the presented object given the natural reduction of non-useful information like background and other parts that might be common in multiple objects (for \eg tires are common to most automobiles and therefore not an important feature in distinguishing between a bike and a motorbike). Such robust representations can help prevent block-box adversarial attacks by attending to only the relevant features in the learned representations and ignoring the rest. 

To this end, Extreme Image Transformations (EIT)~\cite{eit1, eit2} help by selecting variably sized blocks or segments and moving them around, thus breaking the spatial correlations between the foreground and background. Contrary to the MIRCs, which focus on only the specific parts of the object in the image, EITs do not explicitly discard the background, only roughly separate it from the object. This gives them a dual advantage of focusing on the relevant parts of the image and keeping the context intact for better generalizability. More on EITs in \S\ref{subsec:eits_robust_rep_description}.

\paragraph{Contributions} We demonstrate the improvement in performance of neural networks using EITs
\begin{itemize}
    \item We show that off-the-shelf neural networks can be trained on input transformed using Extreme Image Transformations.
    \item We show that EITs alter the internal latent representations of Imagenet-pretrained networks and help with robust learning useful for object recognition.
    \item We show that these weight alterations are linearly dependent on the training regime and training with EITs has effects on all layers of the network.
    \item We show that EIT trained networks better capture the shape of the object and show higher activations in the foreground, suppressing the background. 
    \item We show that EIT trained network performance is generalizable and is not affected as much when tested with higher noise levels.
\end{itemize}

% %%%%%%%%%%%%%%%%%%%%%%
% RESNET50 SALIENCY MAPS
% %%%%%%%%%%%%%%%%%%%%%%

\begin{figure}
    \centering
    \includegraphics{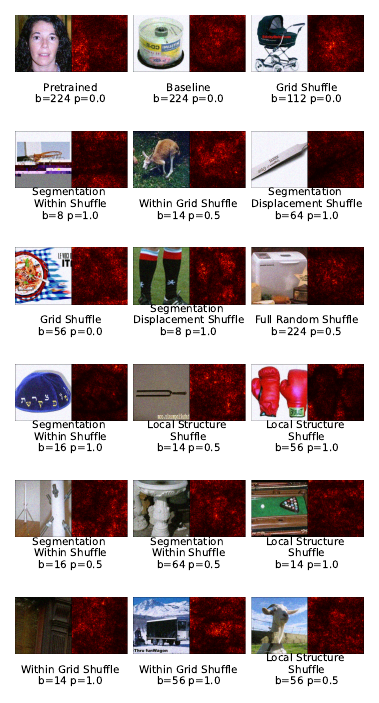}
    \caption{Gaussian Noise level 1 processed test images and saliency maps from Imagenet pretrained ResNet50 used as feature extractor and trained on EITs. The head is trained with the indicated EITs and their hyperparameters on Caltech256 dataset and evaluated on a held-out set. The saliency maps show that EIT trained networks are able to identify the object despite the added gaussian noise on the test dataset.}
    \label{fig:saliency_resnet50_fe_gn1}
\end{figure}

\begin{figure}
    \centering
    \includegraphics{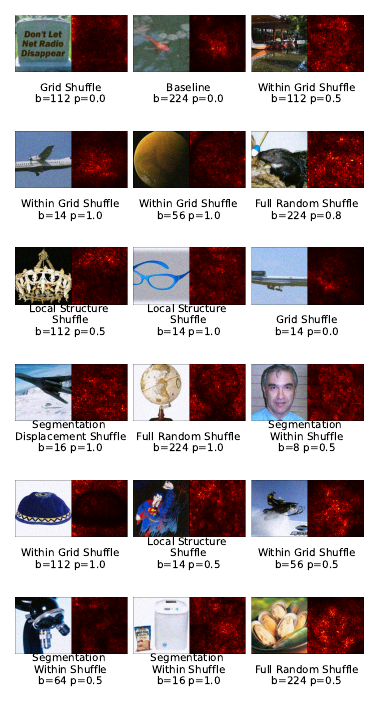}
    \caption{Gaussian Noise level 1 processed test images and saliency maps from Imagenet pretrained ResNet50 finetuned on EITs. The entire network is trained with the indicated EITs and their hyperparameters on Caltech256 dataset and evaluated on a held-out set. The saliency maps show that EIT trained networks are able to identify the object despite the added gaussian noise on the test dataset.}
    \label{fig:saliency_resnet50_ft_gn1}
\end{figure}

\begin{figure}
    \centering
    \includegraphics{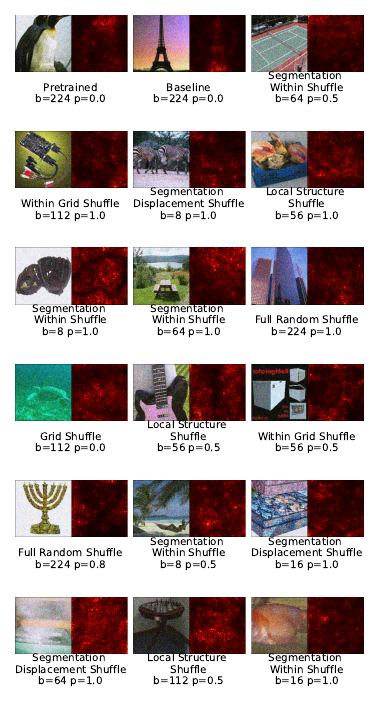}
    \caption{Gaussian Noise level 3 processed test images and saliency maps from Imagenet pretrained ResNet50 used as feature extractor and trained on EITs. The head is trained with the indicated EITs and their hyperparameters on Caltech256 dataset and evaluated on a held-out set. The saliency maps show that EIT trained networks are able to identify the object despite the added gaussian noise on the test dataset.}
    \label{fig:saliency_resnet50_fe_gn3}
\end{figure}

\begin{figure}
    \centering
    \includegraphics{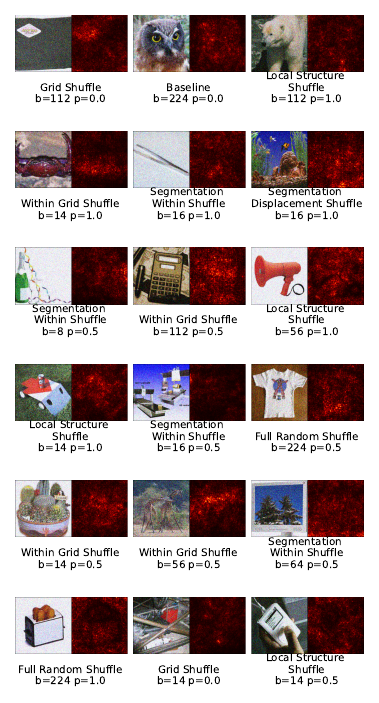}
    \caption{Gaussian Noise level 3 processed test images and saliency maps from Imagenet pretrained ResNet50 finetuned on EITs. The entire network is trained with the indicated EITs and their hyperparameters on Caltech256 dataset and evaluated on a held-out set. The saliency maps show that EIT trained networks are able to identify the object despite the added gaussian noise on the test dataset.}
    \label{fig:saliency_resnet50_ft_gn3}
\end{figure}

\begin{figure}
    \centering
    \includegraphics{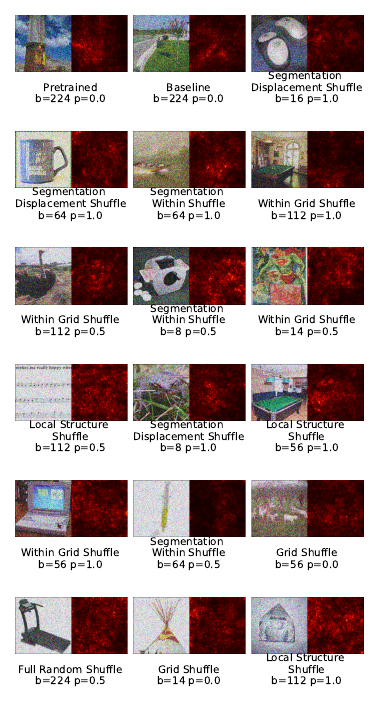}
    \caption{Gaussian Noise level 5 processed test images and saliency maps from Imagenet pretrained ResNet50 used as feature extractor and trained on EITs. The head is trained with the indicated EITs and their hyperparameters on Caltech256 dataset and evaluated on a held-out set. The saliency maps show that EIT trained networks are able to identify the object despite the added gaussian noise on the test dataset.}
    \label{fig:saliency_resnet50_fe_gn5}
\end{figure}

\begin{figure}
    \centering
    \includegraphics{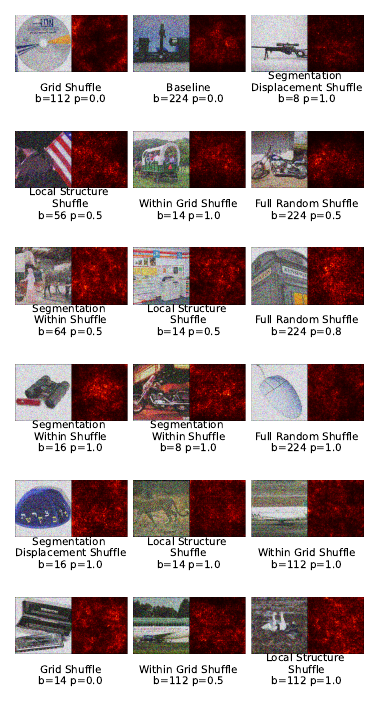}
    \caption{Gaussian Noise level 5 processed test images and saliency maps from Imagenet pretrained ResNet50 finetuned on EITs. The entire network is trained with the indicated EITs and their hyperparameters on Caltech256 dataset and evaluated on a held-out set. The saliency maps show that EIT trained networks are able to identify the object despite the added gaussian noise on the test dataset.}
    \label{fig:saliency_resnet50_ft_gn5}
\end{figure}

% %%%%%%%%%%%%%%%%%%%%%%
% EFFICIENTNETB0 SALIENCY MAPS
% %%%%%%%%%%%%%%%%%%%%%%

\begin{figure}
    \centering
    \includegraphics{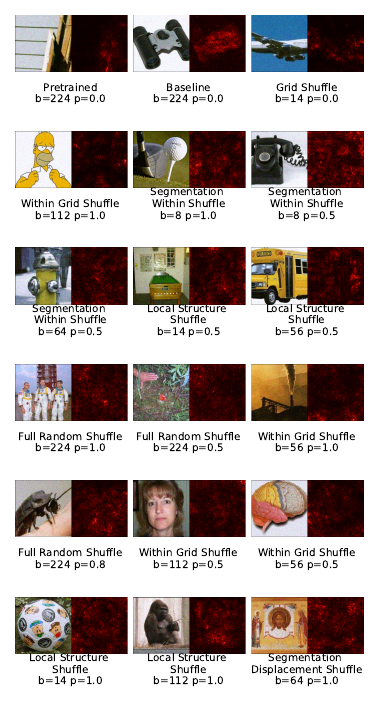}
    \caption{Gaussian Noise level 1 processed test images and saliency maps from Imagenet pretrained EfficientNetB0 used as feature extractor and trained on EITs. The head is trained with the indicated EITs and their hyperparameters on Caltech256 dataset and evaluated on a held-out set. The saliency maps show that EIT trained networks are able to identify the object despite the added gaussian noise on the test dataset.}
    \label{fig:saliency_efficientnetb0_fe_gn1}
\end{figure}

\begin{figure}
    \centering
    \includegraphics{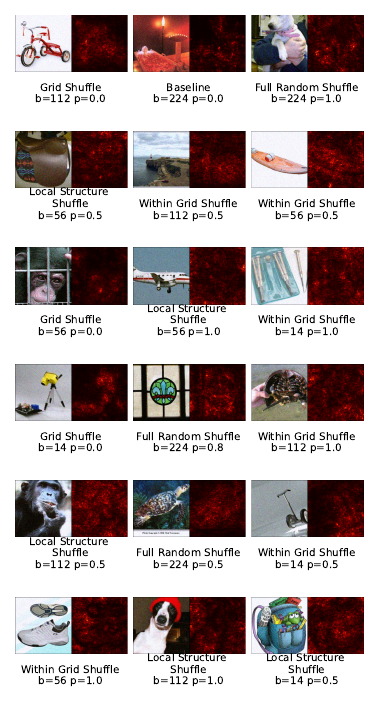}
    \caption{Gaussian Noise level 1 processed test images and saliency maps from Imagenet pretrained EfficientNetB0 finetuned on EITs. The entire network is trained with the indicated EITs and their hyperparameters on Caltech256 dataset and evaluated on a held-out set. The saliency maps show that EIT trained networks are able to identify the object despite the added gaussian noise on the test dataset.}
    \label{fig:saliency_efficientnetb0_ft_gn1}
\end{figure}

\begin{figure}
    \centering
    \includegraphics{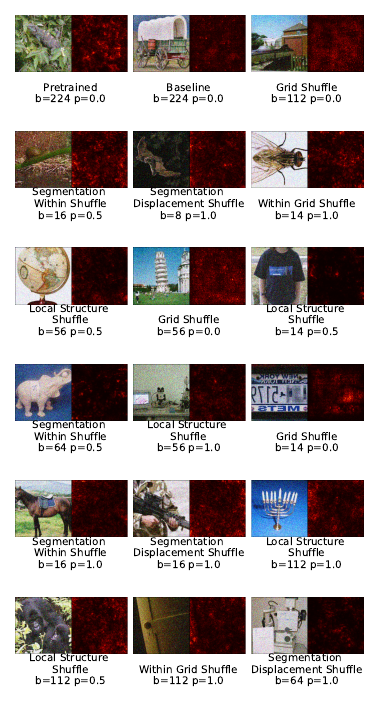}
    \caption{Gaussian Noise level 3 processed test images and saliency maps from Imagenet pretrained EfficientNetB0 used as feature extractor and trained on EITs. The head is trained with the indicated EITs and their hyperparameters on Caltech256 dataset and evaluated on a held-out set. The saliency maps show that EIT trained networks are able to identify the object despite the added gaussian noise on the test dataset.}
    \label{fig:saliency_efficientnetb0_fe_gn3}
\end{figure}

\begin{figure}
    \centering
    \includegraphics{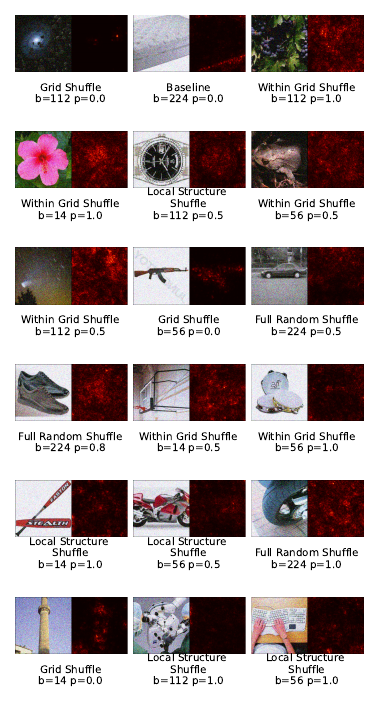}
    \caption{Gaussian Noise level 3 processed test images and saliency maps from Imagenet pretrained EfficientNetB0 finetuned on EITs. The entire network is trained with the indicated EITs and their hyperparameters on Caltech256 dataset and evaluated on a held-out set. The saliency maps show that EIT trained networks are able to identify the object despite the added gaussian noise on the test dataset.}
    \label{fig:saliency_efficientnetb0_ft_gn3}
\end{figure}

\begin{figure}
    \centering
    \includegraphics{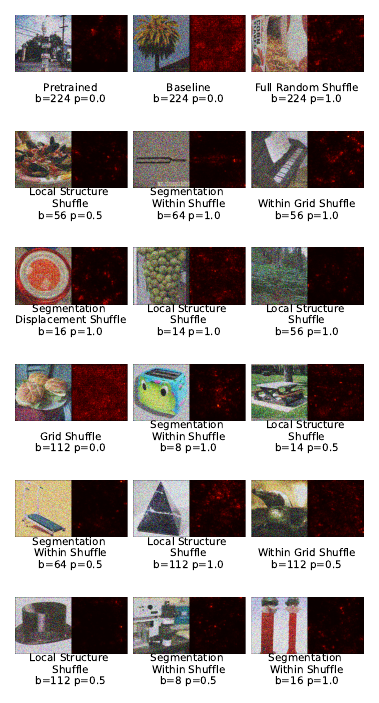}
    \caption{Gaussian Noise level 5 processed test images and saliency maps from Imagenet pretrained EfficientNetB0 used as feature extractor and trained on EITs. The head is trained with the indicated EITs and their hyperparameters on Caltech256 dataset and evaluated on a held-out set. The saliency maps show that EIT trained networks are able to identify the object despite the added gaussian noise on the test dataset.}
    \label{fig:saliency_efficientnetb0_fe_gn5}
\end{figure}

\begin{figure}
    \centering
    \includegraphics{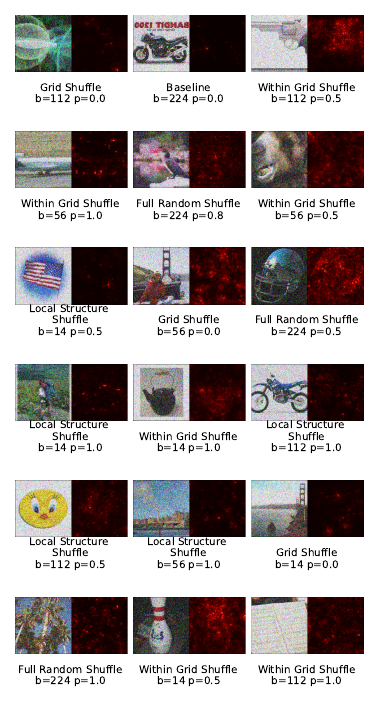}
    \caption{Gaussian Noise level 5 processed test images and saliency maps from Imagenet pretrained EfficientNetB0 finetuned on EITs. The entire network is trained with the indicated EITs and their hyperparameters on Caltech256 dataset and evaluated on a held-out set. The saliency maps show that EIT trained networks are able to identify the object despite the added gaussian noise on the test dataset.}
    \label{fig:saliency_efficientnetb0_ft_gn5}
\end{figure}

% %%%%%%%%%%%%%%%%%%%%%%
% END SALIENCY MAPS
% %%%%%%%%%%%%%%%%%%%%%%

% %%%%%%%%%%%%%%%%%%%%%%
% RESNET50 TABLES
% %%%%%%%%%%%%%%%%%%%%%%
\begin{table}[ht]
    \caption{Test accuracy for ResNet50 network used as a feature extractor. The network was trained on Caltech256 dataset with Block transforms of EIT and validated on gaussian noise level 3 transformed images. The network was tested on a held-out Caltech256 dataset processed with Gaussian Noise levels 1, 3 and 5 indicated as GN1, GN3, and GN5 respectively.}
    \label{tab:resnet_fe_block}
    \begin{tabular}{lcccccc}
    % \begin{tabular}{\linewidth}{@{}*{6}{C}c@{}}
    % \toprule
    \cmidrule[1pt]{1-6}
    
    \multirow{2}{*}{\textbf{Transform}}  & \multirow{2}{*}{\textbf{P}} & \multirow{2}{2em}{\textbf{Grid Size}} & \multicolumn{4}{c}{\textbf{Accuracy (in \%)}}\\  \cmidrule{4-6}
    & & &\textbf{GN 1} & \textbf{GN 3} & \textbf{GN 5}    \\  
    % \midrule
    \cmidrule[1pt]{1-6}

    \text{Pretrained}        &           &           & 0.51 &	 0.27 & 	0.49    \\ \cmidrule{1-6}
    \text{Baseline}             &           &           & 70.59 &	41.25 &	3.66    \\ \cmidrule{1-6}
    % \text{Full Random Shuffle} 
    \multicolumn{1}{c}{\multirow{2}{5em}{Full Random Shuffle}}         & 0.5       &           & 11.40 &	21.68 &	14.41   \\ 
                                & 0.8       &           & 2.99 &	4.09 &	3.37    \\ 
                                & 1.0       &           & 2.79 &	2.84 &	1.45    \\ \cmidrule{1-6}
    \text{Grid Shuffle}         &           & 14x14     & 2.97 &	2.88 &	2.75    \\
                                &           & 56x56     & 28.47 &	15.64 &	3.53    \\ 
                                &           & 112x112     & 66.28 &	40.07 &	3.58    \\ \cmidrule{1-6}
    % \text{Within Grid Shuffle}  
    \multicolumn{1}{c}{\multirow{2}{5em}{Within Grid Shuffle}} & 0.5       & 14x14     & 44.09 &	38.77 &	19.46   \\
                                &           & 56x56     & 17.23 &	18.15 &	13.90   \\
                                &           & 112x112     & 17.18 &	21.21 &	15.51   \\ \cmidrule{2-6} 
                                & 1.0       & 14x14     & 7.06 &	5.43 &	3.13    \\
                                &           & 56x56     & 6.77 &	6.91 &	4.96    \\ 
                                &           & 112x112     & 3.06 &	3.96 &	2.86    \\ \cmidrule{1-6}
    % \text{Local Structure Shuffle}   
    \multicolumn{1}{c}{\multirow{2}{5em}{Local Structure Shuffle}} & 0.5       & 14x14     & 2.86 &	2.79 &	2.77    \\
                                &           & 56x56     & 6.64 &	9.05 &	6.30    \\ 
                                &           & 112x112     & 9.97 &	17.36 &	13.34   \\ \cmidrule{2-6}  
                                & 1.0       & 14x14     & 2.88 &	2.79 &	2.77    \\
                                &           & 56x56     & 4.18 &	4.63 &	3.71    \\ 
                                &           & 112x112     & 2.77 &	3.37 &	2.73    \\ 
    % \bottomrule
    \cmidrule[1pt]{1-6}

    \end{tabular}
    \end{table}

\begin{table}[ht]
\caption{Test accuracy for ResNet50 network used as a feature extractor. The network was trained on Caltech256 dataset with Segmentation transforms of EIT and validated on gaussian noise level 3 transformed images. The network was tested on a held-out Caltech256 dataset processed with Gaussian Noise levels 1, 3 and 5 indicated as GN1, GN3, and GN5 respectively.}
\label{tab:resnet_fe_seg}
\begin{tabular}{lcccccc}
% \toprule
\cmidrule[1pt]{1-6}

\multirow{2}{*}{\textbf{Transform}}  & \multirow{2}{*}{\textbf{P}} & \multirow{2}{2em}{\textbf{Grid Size}} & \multicolumn{4}{c}{\textbf{Accuracy (in \%)}}\\  \cmidrule{4-6}
& & &\textbf{GN 1} & \textbf{GN 3} & \textbf{GN 5}    \\  
% \midrule
\cmidrule[1pt]{1-6}

\multicolumn{1}{c}{\multirow{2}{8em}{Segmentation Displacement Shuffle}}  &      & 8        & 9.70 &	8.47 &	6.19 \\
                             &           & 16       & 9.81 &	9.94 &	7.78 \\ 
                             &           & 64       & 7.60 &	6.99 &	5.43 \\ \cmidrule{1-6}
\multicolumn{1}{c}{\multirow{2}{8em}{Segmentation Within Shuffle}}              & 0.5       & 8        & 11.93 &	11.71 &	10.84 \\
                             &           & 16       & 13.97 &	13.63 &	11.78 \\ 
                             &           & 64       & 17.18 &	16.16 &	11.62 \\ \cmidrule{2-6} 
                             & 1.0       & 8        & 6.84 &	7.22 &	6.28 \\
                             &           & 16       & 7.69 &	7.96 &	6.86 \\ 
                             &           & 64       & 8.65 &	8.02 &	5.90 \\ 
% \bottomrule
\cmidrule[1pt]{1-6}

\end{tabular}
\end{table}

\begin{table}[ht]
    \caption{Test accuracy for finetuned ResNet50 network. The entire network was finetuned on Caltech256 dataset with Block transforms of EIT and validated on gaussian noise level 3 transformed images. The network was tested on a held-out Caltech256 dataset processed with Gaussian Noise levels 1, 3 and 5 indicated as GN1, GN3, and GN5 respectively.}
    \label{tab:resnet_ft_block}
    \begin{tabular}{lcccccc}
    \cmidrule[1pt]{1-6}
    
    \multirow{2}{*}{\textbf{Transform}}  & \multirow{2}{*}{\textbf{P}} & \multirow{2}{2em}{\textbf{Grid Size}} & \multicolumn{4}{c}{\textbf{Accuracy (in \%)}}\\  \cmidrule{4-6}
    & & &\textbf{GN 1} & \textbf{GN 3} & \textbf{GN 5}    \\  
    \cmidrule[1pt]{1-6}

    \text{Pretrained}        &           &           & 0.51 &	 0.27 & 	0.49    \\ \cmidrule{1-6}
    \text{Baseline}             &           &           & 74.53 &	47.31 &	6.73    \\ \cmidrule{1-6}
    \multicolumn{1}{c}{\multirow{2}{5em}{Full Random Shuffle}}         & 0.5       &           & 72.65 &	69.83 &	47.98   \\ 
                                & 0.8       &           & 32.80 &	32.42 &	21.52   \\ 
                                & 1.0       &           & 2.70 &	2.30 &	2.77    \\ \cmidrule{1-6}
    \text{Grid Shuffle}         &           & 14x14     & 15.08 &	8.69 &	1.39    \\
                                &           & 56x56     & 66.12 &	37.23 &	4.69    \\ 
                                &           & 112x112     & 73.14 &	44.98 &	7.20    \\ \cmidrule{1-6}
    \multicolumn{1}{c}{\multirow{2}{5em}{Within Grid Shuffle}} & 0.5       & 14x14     & 68.13 &	45.94 &	13.47   \\
                                &           & 56x56     & 72.76 &	64.51 &	39.20   \\
                                &           & 112x112     & 73.94 &	69.85 &	46.66   \\ \cmidrule{2-6} 
                                & 1.0       & 14x14     & 40.29 &	16.18 &	4.34    \\
                                &           & 56x56     & 5.56 &	4.38 &	3.53    \\ 
                                &           & 112x112     & 2.59 &	3.15 &	2.64    \\ \cmidrule{1-6}
    \multicolumn{1}{c}{\multirow{2}{5em}{Local Structure Shuffle}} & 0.5       & 14x14     & 4.69 &	3.46 &	2.99    \\
                                &           & 56x56     & 48.25 &	41.50 &	21.72   \\ 
                                &           & 112x112     & 70.32 &	65.50 &	44.29   \\ \cmidrule{2-6}  
                                & 1.0       & 14x14     & 4.18 &	3.15 &	2.15    \\
                                &           & 56x56     & 4.09 &	3.93 &	3.06    \\ 
                                &           & 112x112     & 2.70 &	3.20 &	2.57    \\ 
    \cmidrule[1pt]{1-6}

    \end{tabular}
    \end{table}

\begin{table}[ht]
    \caption{Test accuracy for finetuned ResNet50 network. The entire network was finetuned on Caltech256 dataset with Segmentation transforms of EIT and validated on gaussian noise level 3 transformed images. The network was tested on a held-out Caltech256 dataset processed with Gaussian Noise levels 1, 3 and 5 indicated as GN1, GN3, and GN5 respectively. The experiments that could not be excuted due to resource constraints are indicated by T/O.}
    \label{tab:resnet_ft_seg}
    \begin{tabular}{lcccccc}
    % \toprule
    \cmidrule[1pt]{1-6}
    
    \multirow{2}{*}{\textbf{Transform}}  & \multirow{2}{*}{\textbf{P}} & \multirow{2}{2em}{\textbf{Grid Size}} & \multicolumn{4}{c}{\textbf{Accuracy (in \%)}}\\  \cmidrule{4-6}
    & & &\textbf{GN 1} & \textbf{GN 3} & \textbf{GN 5}    \\  
    % \midrule
    \cmidrule[1pt]{1-6}
    
    \multicolumn{1}{c}{\multirow{2}{8em}{Segmentation Displacement Shuffle}}  &      & 8        & 31.31 &	14.28 &	4.72    \\
                                 &           & 16       & 27.78 &	12.76 &	4.67    \\ 
                                 &           & 64       & 27.55 &	12.29 &	3.46    \\ \cmidrule{1-6}
    \multicolumn{1}{c}{\multirow{2}{8em}{Segmentation Within Shuffle}}              & 0.5       & 8        & 18.61 &	15.96 &	10.46   \\
                                 &           & 16       & 15.75 &	13.77 &	10.68   \\ 
                                 &           & 64       & 9.99 &	9.45 &	8.54    \\ \cmidrule{2-6} 
                                 & 1.0       & 8        & 35.22 &	15.53 &	3.80    \\
                                 &           & 16       & 5.81 &	5.01 &	4.09    \\ 
                                 &           & 64       & T/O & T/O & T/O \\ 
    % \bottomrule
    \cmidrule[1pt]{1-6}
    
    \end{tabular}
    \end{table}

% %%%%%%%%%%%%%%%%%%%%%%
% EFFICIENTNETB0 TABLES
% %%%%%%%%%%%%%%%%%%%%%%
\begin{table}[ht]
    \caption{Test accuracy for EfficientNetB0 network used as a feature extractor. The network was trained on Caltech256 dataset with Block transforms of EIT and validated on gaussian noise level 3 transformed images. The network was tested on a held-out Caltech256 dataset processed with Gaussian Noise levels 1, 3 and 5 indicated as GN1, GN3, and GN5 respectively.}
    \label{tab:effnet_fe_block}
    \begin{tabular}{lcccccc}
    \cmidrule[1pt]{1-6}
    
    \multirow{2}{*}{\textbf{Transform}}  & \multirow{2}{*}{\textbf{P}} & \multirow{2}{2em}{\textbf{Grid Size}} & \multicolumn{4}{c}{\textbf{Accuracy (in \%)}}\\  \cmidrule{4-6}
    & & &\textbf{GN 1} & \textbf{GN 3} & \textbf{GN 5}    \\  
    \cmidrule[1pt]{1-6}

    \text{Pretrained}        &           &           & 0.49 &	0.25 &	0.34    \\ \cmidrule{1-6}
    \text{Baseline}             &           &           & 2.01 &	3.02 &	3.06    \\ \cmidrule{1-6}
    \multicolumn{1}{c}{\multirow{2}{5em}{Full Random Shuffle}}         & 0.5       &           & 5.72 &	4.20 &	2.48    \\ 
                                & 0.8       &           & 4.00 &	4.25 &	3.11    \\ 
                                & 1.0       &           & 3.84 &	3.37 &	2.79    \\ \cmidrule{1-6}
    \text{Grid Shuffle}         &           & 14x14     & 2.97 &	2.32 &	2.48    \\
                                &           & 56x56     & 1.97 &	2.53 &	3.11    \\ 
                                &           & 112x112     & 2.23 &	2.59 &	1.92    \\ \cmidrule{1-6}
    \multicolumn{1}{c}{\multirow{2}{5em}{Within Grid Shuffle}} & 0.5       & 14x14     & 13.54 &	5.74 &	2.57    \\
                                &           & 56x56     & 9.27 &	5.32 &	1.92    \\
                                &           & 112x112     & 8.36 &	4.89 &	3.24    \\ \cmidrule{2-6} 
                                & 1.0       & 14x14     & 4.76 &	3.28 &	1.50    \\
                                &           & 56x56     & 2.66 &	1.99 &	0.76    \\ 
                                &           & 112x112     & 2.86 &	1.90 &	2.68    \\ \cmidrule{1-6}
    \multicolumn{1}{c}{\multirow{2}{5em}{Local Structure Shuffle}} & 0.5       & 14x14     & 3.75 &	2.46 &	0.63    \\
                                &           & 56x56     & 4.94 &	3.13 &	1.30    \\ 
                                &           & 112x112     & 5.94 &	4.00 &	2.84    \\ \cmidrule{2-6}  
                                & 1.0       & 14x14     & 3.60 &	2.73 &	0.58    \\
                                &           & 56x56     & 2.77 &	2.64 &	1.85    \\ 
                                &           & 112x112     & 2.82 &	1.65 &	2.17    \\ 
    \cmidrule[1pt]{1-6}

    \end{tabular}
    \end{table}

\begin{table}[ht]
\caption{Test accuracy for EfficientNetB0 network used as a feature extractor. The network was trained on Caltech256 dataset with Segmentation transforms of EIT and validated on gaussian noise level 3 transformed images. The network was tested on a held-out Caltech256 dataset processed with Gaussian Noise levels 1, 3 and 5 indicated as GN1, GN3, and GN5 respectively.}
\label{tab:effnet_fe_seg}
\begin{tabular}{lcccccc}
\cmidrule[1pt]{1-6}

\multirow{2}{*}{\textbf{Transform}}  & \multirow{2}{*}{\textbf{P}} & \multirow{2}{2em}{\textbf{Grid Size}} & \multicolumn{4}{c}{\textbf{Accuracy (in \%)}}\\  \cmidrule{4-6}
& & &\textbf{GN 1} & \textbf{GN 3} & \textbf{GN 5}    \\  
\cmidrule[1pt]{1-6}

\multicolumn{1}{c}{\multirow{2}{8em}{Segmentation Displacement Shuffle}}  &      & 8        & 4.13 &	3.91 &	0.45    \\
                             &           & 16       & 4.36 &	3.82 &	0.49    \\ 
                             &           & 64       & 6.46 &	2.99 &	0.69    \\ \cmidrule{1-6}
\multicolumn{1}{c}{\multirow{2}{8em}{Segmentation Within Shuffle}}              & 0.5       & 8        & 7.13 &	5.72 &	0.42    \\
                             &           & 16       & 8.45 &	6.66 &	0.31    \\ 
                             &           & 64       & 12.56 &	7.08 &	0.51    \\ \cmidrule{2-6} 
                             & 1.0       & 8        & 2.59 &	2.41 &	0.60    \\
                             &           & 16       & 2.77 &	2.82 &	0.22    \\ 
                             &           & 64       & 4.47 &	2.39 &	0.31    \\ 
\cmidrule[1pt]{1-6}

\end{tabular}
\end{table}

\begin{table}[ht]
    \caption{Test accuracy for finetuned EfficientNetB0 network. The entire network was finetuned on Caltech256 dataset with Block transforms of EIT and validated on gaussian noise level 3 transformed images. The network was tested on a held-out Caltech256 dataset processed with Gaussian Noise levels 1, 3 and 5 indicated as GN1, GN3, and GN5 respectively.}
    \label{tab:effnet_ft_block}
    \begin{tabular}{lcccccc}
    \cmidrule[1pt]{1-6}
    
    \multirow{2}{*}{\textbf{Transform}}  & \multirow{2}{*}{\textbf{P}} & \multirow{2}{2em}{\textbf{Grid Size}} & \multicolumn{4}{c}{\textbf{Accuracy (in \%)}}\\  \cmidrule{4-6}
    & & &\textbf{GN 1} & \textbf{GN 3} & \textbf{GN 5}    \\  
    \cmidrule[1pt]{1-6}

    \text{Pretrained}        &           &           & 0.49 &	0.25 &	0.34    \\ \cmidrule{1-6}
    \text{Baseline}             &           &           & 52.20 &	4.69 &	2.57    \\ \cmidrule{1-6}
    \multicolumn{1}{c}{\multirow{2}{5em}{Full Random Shuffle}}         & 0.5       &           & 52.34 &	60.02 &	47.40   \\ 
                                & 0.8       &           & 6.35 &	7.66 &	6.10    \\ 
                                & 1.0       &           & 2.77 &	3.20 &	2.66    \\ \cmidrule{1-6}
    \text{Grid Shuffle}         &           & 14x14     & 16.60 &	3.08 &	0.65    \\
                                &           & 56x56     & 49.68 &	9.10 &	2.82    \\ 
                                &           & 112x112     & 53.72 &	9.92 &	3.15    \\ \cmidrule{1-6}
    \multicolumn{1}{c}{\multirow{2}{5em}{Within Grid Shuffle}} & 0.5       & 14x14     & 63.37 &	33.39 &	8.65    \\
                                &           & 56x56     & 61.07 &	60.87 &	36.22   \\
                                &           & 112x112     & 64.87 &	67.13 &	57.74   \\ \cmidrule{2-6} 
                                & 1.0       & 14x14     & 27.62 &	11.66 &	4.34    \\
                                &           & 56x56     & 4.07 &	3.51 &	2.75    \\ 
                                &           & 112x112     & 2.41 &	1.97 &	1.41    \\ \cmidrule{1-6}
    \multicolumn{1}{c}{\multirow{2}{5em}{Local Structure Shuffle}} & 0.5       & 14x14     & 3.11 &	3.33 &	1.18    \\
                                &           & 56x56     & 26.88 &	22.44 &	11.13   \\ 
                                &           & 112x112     & 42.77 &	45.23 &	29.14   \\ \cmidrule{2-6}  
                                & 1.0       & 14x14     & 3.37 &	2.88 &	1.92    \\
                                &           & 56x56     & 3.46 &	2.93 &	3.55    \\ 
                                &           & 112x112     & 1.59 &	0.96 &	1.03    \\ 
    \cmidrule[1pt]{1-6}

    \end{tabular}
    \end{table}
 
% %%%%%%%%%%%%%%%%%%%%%%
% END RESULT TABLES
% %%%%%%%%%%%%%%%%%%%%%%

\section{Related Works}

\paragraph{Adversarially trained models}
Unnoticeable perturbations to the ``minimal recognizable images" can affect the network performance~\cite{ullman2016atoms}. Altering texture information and silhouette contours has shown to help with robust performance against diverse image corruptions~\cite{baker2018deep}. Rusak et al. show that adversarial training with locally correlated noise has shown to improve performance by altering spatial biases~\cite{rusak2020simple}. Baradad et al.~\cite{learningtoseebylookingatnoise} try to learn robust visual representations by generating models of noise closer to the distribution of real images. Adding noise to training inputs has been considered to be a form of $L_2$ regularization~\cite{bishop1995training, cohen2019certified}. However, adversarial training is sensitive to the simple parameters like weight decay, a change in which could result in significant performance degradation~\cite{pang2020bag}. Madry et al. suggest that increasing the capacity of networks could help with robust perfomance against adversarial attacks~\cite{madry2017towards}. Other studies have proposed unsupervised label-free learning to improve adversarial robustness~\cite{alayrac2019labels} and used ensemble training with uncorrelated loss functions~\cite{kariyappa2019improving}.

\paragraph{Adding layers to alter statistics}
Multiple network layers have been proposed to help with robustness against adversarial attacks. Xie et al.~\cite{xie2019feature} propose feature denoising on the adversarially attacked images, while Chen et al.~\cite{chen2021robust} propose that separate batchnorm layers be used to maintain distinct statistics for clean and adversarial training images, called Det-AdvProp, which is an improvement of the original AdvProp~\cite{xie2020adversarial}. Shu et al.~\cite{shu2021encoding} propose Adversarial Batch Normalization layer to be added to a network that can shift feature statistics and re-normalize them to the most damaging mean and variance right before a gradient update. 

% \paragraph{Training with noise transformed input} -- covered in adv trained models

\paragraph{Training with objects}
% training with object silhouettes/contours and breaking the input into squares/other shapes 
Multiple studies have altered the input to more directly represent the object being learned by using shapes as inputs or replacing the texture of the object with another object, and hence introducing a shape bias~\cite{geirhos2018generalisation, geirhos2018imagenet}, or learning from shapes directly~\cite{felzenszwalb2001learning}. Tolstikhin et al.~\cite{tolstikhin2021mlpmixer} show the applying using multi-layer perceptrons on image cropped into patches and then connecting them with another multi-layer perceptron can reach or surpass Vision Transformers (ViT) performance~\cite{dosovitskiy2021an}. Not only humans but birds also rely on edges and contours of object during recognition~\cite{peissig2005role}.

\subsection{Extreme Image Transformations (EIT)}
\label{subsec:eits_robust_rep_description}
Humans can make near accurate decisions based on a very limited set of features, demonstrating a natural dimensionality reduction in V1~\cite{pang2016dimensionality, cocci2015cortical, wang2014dimensional, sarti2015constitution}. Previous studies have found that seemingly insignificant changes in visual input adversarially impact network performance but do not influence human performance. Humans are able to reliably detect objects up to a minimum recognizable configuration (MIRC) ---- the smallest image dimensions centered at the most important features of the object~\cite{ullman2016atoms}. Extreme Image Transforms probe this observation to see if networks can learn reliable representations when trained on parametrically shuffled visual input~\cite{eit1, eit2}. EITs are a set of seven unique transforms inspired by neurophysiological findings (\textit{Full Random Shuffle, Grid Shuffle, Within Grid Shuffle, Local Structure Shuffle, Color Flatten, Segmentation Displacement Shuffle} and \textit{Segmentation Within Shuffle}). These transforms shuffle the input image based on combination of 3 parameters ---- size of the grid/number of segments, probability of pixel shuffling, and a binary grid/segment swap. We use the block and segmentation transforms as defined in~\cite{eit2}, except color flatten given computational constraints. 

\section{Experiments}
We used ImageNet~\cite{imagenet} pretrained ResNet50~\cite{resnet} and EfficientNetB0~\cite{efficientnet} networks with Extreme Image Transformations (EIT)~\cite{eit1, eit2} applied to Caltech256 dataset~\cite{caltech256}. ResNet50 is perhaps the simplest model with top scores on BrainScore~\cite{brainscore} and usually forms the backbone for a lot of models in the adversarial attack and object recognition literature. We chose EfficientNetB0 for its high overall average score on BrainScore~\cite{brainscore} at the time of experiments and also its low computational footprint. EfficientNets also demonstrated strong performance against adversarial attacks with minimal changes in recording and maintaining batchnorm parameters~\cite{xie2020adversarial}. We use these models as feature extractors and also finetune them end-to-end using EITs to probe the capabilities of EITs in learning robust object representations. 

% what networks did you pick - resnet50 and edfficientnetb0, and why (quick to train, resnet is backbone for many networks, efficientnet has had successes with Adaprop layers and has a good brainscore, also cheapest to train among the 8 variants.)?

% what EITs did you use? What hyperparams and why (image size to be divided into 1/4th and so on)? (mention about not being able to use the segmentation EITs for computational reasons.)

\subsection{Dataset and Preprocessing}
\label{subsec:dataset}
% talk about processing Caltech256 with EITs first use as training dataset, and then use them processed with Imagenet-C corruption as validation set. 
% how was caltech256 split, how many images are in each split.

% also talk about the training regime and practices -- early stopping, batch size, lr, momentum, etc. 
We used Caltech256 dataset~\cite{caltech256} for our experiments given its wide variety of 256 object classes and 1 extra clutter class containing images that cannot be classified into any other category. The clutter class can be treated as miscellaneous, with variation in image categories ranging from objects to abstract patterns. The object categories in Caltech256 are more coarse-grained than Imagenet~\cite{imagenet}; \eg Imagenet categorises a particular breed of an animal into one class and a different breed of the same animal into another class, while Caltech256 puts all different breeds of the animal into a single class. 

We segregated Caltech256 ($30,607$ images) into three partitions --- train ($21,657$ images), validation ($4,475$ images) and test ($4,475$ images) sets. We transformed the training set with our EITs and their hyperparameters~\cite{eit1, eit2}, while the validation and test sets were transformed with gaussian noise of severities 1, 3 and 5 as stated in \S\ref{subsec:eit_featureextraction}, \ref{subsec:eit_finetuning}. The gaussian noise corruption is generated in the same way as in the Imagenet-C dataset~\cite{imagenet-c}. 

For all our experiments we applied early stopping if the validation loss did not improve for 200 epochs straight. For feature extraction experiments, we used a batch size of $1024$ with both ResNet50 and EfficientNetB0. For finetuning, the batch size was reduced to $256$ for both ResNet50 and EfficientNetB0. We used a learning rate of $10^{-3}$ and a momemntum of $0.9$ for all our experiments. The inference for both feature extraction and finetuning experiments was run with a batch size of $1024$ across both networks. We used stochastic gradient descent optimizer. We also reduced the learning rate by a factor of $0.1$ when the optimizer did not show a minimization in loss for 10 straight epochs.

% what was the baseline (no train/val data, just testing pretrained weights with caltech256), what were the primary experiments with EITs and no transform. 

% talk about the preliminary tests with highest val score in gaussian noise. Also this is most prone to happening in the real world (needs citation)
\paragraph{Why Gaussian noise corruption?} Given over 18 different kinds of corruptions in Imagenet-C and over 27 unique combinations of EITs, it was computationally challenging to evaluate the entire matrix with about 500 experiments, for each network and their training processes. Based on literature~\cite{brainscore} and computational demands, we found ResNet50 to be the most representative of the networks. We used an Imagenet-pretrained ResNet50 as a feature extractor and trained only the readout layers with different variations of EITs. We then evaluated the individual models on images corrupted with a level $3$ severity of \textit{fog, gaussian blur, gaussian noise, snow} and \textit{spatter} ---- a set of most representative corruptions found abundantly in the real world settings. We repeated the same process using both Imagenet validation set randomly split into train and validation, and Caltech256 dataset randomly split into train and validation. We found the average and median validation accuracy of gaussian noise corruption to be the clear winner in case of Caltech256 and a close second in case of Imagenet across all variations of trainings with EITs and their hyperparameters.

% We used gaussian noise given its simplicity and widespread use in real-world applications. 

\subsection{Feature extraction}
\label{subsec:eit_featureextraction}
We started with a ResNet50~\cite{resnet} and an EfficientNetB0~\cite{efficientnet} network pretrained on Imagenet-1K~\cite{imagenet} dataset. We froze the weights of all but readout layers in the networks. We then trained the readout layers, starting with the Imagenet weights, on a partition of Caltech256~\cite{caltech256} dataset preprocessed using our EITs. For validation dataset we used another unique partition of Caltech256 dataset preprocessed with gaussian noise of severity level 3 from Imagenet-C corruptions~\cite{imagenet-c}. We tested our readout-finetuned networks with EITs on another unique partition of Caltech256 preprocessed with gaussian noise corruptions of severity levels 1, 3 and 5.

\subsection{Finetuning}
\label{subsec:eit_finetuning}
EITs aim to reorient the latent representations of the networks to learn robust object representations instead of shortcuts~\cite{eit1, eit2, geirhos2020shortcut, luo2021rectifying}. To probe the effectiveness of EITs in strengthening the latent representations of all the layers, we used EITs to finetune the entire network instead of only the readout as done in \S\ref{subsec:eit_featureextraction}. We started with a ResNet50~\cite{resnet} and EfficientNetB0~\cite{efficientnet} network pretrained on Imagenet-1K~\cite{imagenet} dataset. We finetuned the entire network, initialized with Imagenet weights, using a partition of Caltech256 dataset~\cite{caltech256} preprocessed with EITs. Similar to the previous strategy, we used another unique partition of Caltech256 as validation dataset preprocessed with gaussian noise of severity level 3 from the Imagenet-C corruptions~\cite{imagenet-c}. We tested the resulting finetuned networks on a separate partition of Caltech256 preprocessed with gaussian noise corruptions of severity levels 1, 3 and 5. 

\section{Results: EITs help learn robust object representations}
\label{sec:results_eit_robust_obj_rep}
Imagenet pretrained ResNet50 has the lowest accuracy at 0.5\%, 0.2\% and 0.4\% when tested with gaussian noise levels 1, 3 and 5 respectively without any training as a feature extractor or validation with image corruptions. We treat training without transformations and validation with gaussian noise level 3 as baselines. For ResNet50 feature extraction expriments, the test accuracy with gaussian noise levels 1, 3 and 5 is at 70.5\%, 41.2\% and 3.6\% respectively. When used as a feature extractor and trained with Grid Shuffle EIT using a grid size of 112, ResNet50 has a test accuracy of 66.2\% with gaussian noise level 1 and of 40\% with gaussian noise level 3; gaussian noise level 5 has the test accuracy plummeting to 3.5\%. Within Grid Shuffle EIT with a grid size of 14 and shuffle probability of 0.5 yields the highest test accuracy for gaussian noise level 5 at 19.4\%; gaussian noise levels 1 and 3 also are a close second at 44\% and 38.7\% respectively (see Tables~\ref{tab:resnet_fe_block}, \ref{tab:resnet_fe_seg}). 

We saw much higher numbers when the entire ResNet50 network is finetuned. The test accuracy for baselines is 74.5\%, 47.3\% and 6.7\% for test with gaussian noise levels 1, 3 and 5 respectively. Finetuning with EITs and validating with gaussian noise level 3 corruption, we found Within Grid Shuffle EIT with a grid size of 112 and shuffle probability of 0.5 to perform the best at 73.9\% and 69.8\% test accuracy on gaussian noise level 1 and 3 respectively. While gaussian noise level 3 performs at 46.6\% for this EIT, the best performance of 47.9\% on gaussian noise level 5 is achieved by finetuning using Randomized Image Shuffle with a shuffle probability of 0.5. The performance on gaussian noise levels 1 and 3 are also a close second on Randomized Image Shuffle (see Tables~\ref{tab:resnet_ft_block}, \ref{tab:resnet_ft_seg}). 

We also tested EfficientNetB0 in similar experiments as with ResNet50. An Imagenet pretrained EfficientNetB0 also had the lowest accuracy at 0.4\%, 0.2\% and 0.3\% for gaussian noise levels 1, 3 and 5 corruptions. The baselines with EfficientNetB0 being used as feature extractor without any training on EITs and only validation on gaussian noise level 3 corruption yielded 2\%, 3\% and 3\% when tested with gaussian noise level 1, 3 and 5 corruptions respectively. For gaussian noise level 1, training on Within Grid Shuffle with a grid size of 14 and shuffle probability of 0.5 performed the best at 13.5\%. Increasing the grid size to 112 gives the best accuracy of 3.2\% when tested on gaussian noise level 5. For gaussian noise level 3, training on Segmentation Shuffle Within with 64 segments and shuffle probability of 0.5 gives the best accuracy of 7\% (see Tables~\ref{tab:effnet_fe_block}, \ref{tab:effnet_fe_seg}). 

As observed with ResNet50, finetuning entire EfficientNetB0 network also yields better results than EfficientNetB0 used as a feature extractor. The test accuracy for finetuning with baselines and testing on gaussian noise levels 1, 3 and 5 was 52.2\%, 4.6\% and 2.5\% respectively. While the absolute accuracy numbers for EfficientNetB0 finetuning are less than those of ResNet50 finetuning, they are much higher than EfficientNetB0 being used as a feature extractor. We found that finetuning using Within Grid Shuffle EIT with a grid size of 112 and shuffle probability of 0.5 yields the highest accuracy at 64.8\%, 67.1\% and 57.7\% across gaussian noise levels 1, 3 and 5 test sets respectively (see Table~\ref{tab:effnet_ft_block}). Due to computational constrains we could not run the EfficientNetB0 finetuning experiments with segmentation based EITs.  

\subsection{How do EITs facilitate learning?}
Much to the contrary belief in deep learning paradigm of throwing all the data at the algorithm~\cite{sun2017revisiting}, human brain filters features at almost every stage, ranging from shape to contrast to direction selectivity~\cite{luck1994spatial, geisler2001edge, moutoussis2008motion, dupont1994many, culham2001visual, sunaert1999motion}. Previous studies have shown the limit to which humans need to look at the features to describe the image~\cite{ullman2016atoms, hummel2013object, whitney2011visual, tarr1998image, wyatte2012limits}. EITs try to facilitate a similar kind of feature selection by breaking the dependency on spatial arrangement in images and/or exploiting shortcuts. EITs work in a similar manner as other transforms like cropping and normalization applied to input pre-processing. The block-style transforms try to capture the bare-minimum features in the individual blocks that represent the object. It also helps with cutting down on the background to help generalize the latent representation to different conditions and environments, similar to how some networks~\cite{pathtracker, NEURIPS2021linsley} and humans have a selectivity bias~\cite{braddick2007development, driver1995object, grossberg1987neural, grossberg2001neural}. 

% write about what you see in the saliency maps
Using ResNet50 as a feature extractor, we see only see a limited change in the accuracy compared to the pretrained and baseline cases, given weights for only the readout layers were updated. The EIT updated networks show higher activations on the objects tested with a higher level of gaussian noise. We however see that they could still be prone to confusion between the object and the background given higher activations on the background areas instead of the object. This highlights a blindspot in the Imagenet training of these popular models, which are prone to exploiting shortcuts and do not capture the fundamental features of the object~\cite{geirhos2020shortcut, luo2021rectifying}. The lowest performance of pretrained networks being tested with noisy test datasets is also visible through the activation scattered saliency maps. See Figures~\ref{fig:saliency_resnet50_fe_gn1}, \ref{fig:saliency_resnet50_fe_gn3} and \ref{fig:saliency_resnet50_fe_gn5}. 

The improvements are more pronounced when the weights of the entire network are updated with EITs. As seen in Figures~\ref{fig:saliency_resnet50_ft_gn1}, \ref{fig:saliency_resnet50_ft_gn3} and \ref{fig:saliency_resnet50_ft_gn5}, the activations in the saliency maps are more centered on the object and capture not only the exact shape of a single but multiple objects/subparts of the object in picture. There is a considerable suppression of the background clutter even when it resembles the texture of the object at higher noise levels. While simple validation seems to work for baselines, we do see significant increases in accuracy and activations when the networks are partially or fully re-trained with EITs, indicating the alteration of object representations being done by EITs. 

In line with the accuracy numbers shown by the re-training of readout layers in EfficientNetB0 using EITs, the activations in saliency maps are also not very high and are mostly scattered. This could be due to  multiple reasons including not enough time to train, stopped too early, not enough training data compared to the number of parameters in the network, etc. The top performers also have accuracy in low double digits and do not show much change in the weights, as is seen by their similarity to the pretrained case. See Figures~\ref{fig:saliency_efficientnetb0_fe_gn1}, \ref{fig:saliency_efficientnetb0_fe_gn3} and \ref{fig:saliency_efficientnetb0_fe_gn5}. 

The difference is again very patent in case when all weights of the network are changed in re-training with EITs, indicating the capabilities of EITs in helping with object recognition in case of noisy inputs and adversarial attacks. Figures~\ref{fig:saliency_efficientnetb0_ft_gn1}, \ref{fig:saliency_efficientnetb0_ft_gn3} and \ref{fig:saliency_efficientnetb0_ft_gn5} indicate higher activations centered on the object with a suppressed background. Background suppression is stronger than foreground activations compared to ResNet50. One potential limitation that we can see in case of EfficientNetB0 is the inability to generalize backwards to synthetic images of the object classes. This could be analogous to a `real2sim` problem when looking at the latent object representations~\cite{hahn2019real2sim, heiden2019real2sim, chen2022real2sim, rao2020rl}.

\section{Discussion and Conclusion}
Our work is inspired by the idea of uncovering \text{``objectness"} in the visual input being fed to machines to facilitate learning a robust set of features for object recognition. This has been done by using object contours either in isolation or augmented with the input being fed to the deep networks~\cite{geirhos2018generalisation, felzenszwalb2001learning}. Downside of this approach is the computational cost of preprocessing the input dataset and the relatively limited accuracy of segmentation algorithms being used. A few approaches went a step ahead and altered the texture inside the contours to help learn a robust representation of objects~\cite{geirhos2018imagenet}. 

Input transformations are also widely used to prevent impact to performance due to perturbations~\cite{guo2017countering, zhang2021adversarial}. Most transformations used to prevent adversarial attacks are augmentative in nature and add extra information to the input~\cite{shu2021encoding, learningtoseebylookingatnoise, poursaeed2018generative}. EITs are not bound by these information augmentation problems or high computational costs. They implement a simple yet effective technique to transform the inputs based on neurophysiological data~\cite{edelman1997complex, tarr1998image, biederman1991priming, ferrari2007groups,hubel1962receptive, hubel1963shape, wiesel1963single, hubel1963receptive, wiesel1963effects, tanaka1997mechanisms, grill2001lateral}. They have probed the limits of recognition in humans and machines and highlighted the blindspots for popular networks~\cite{eit1, eit2}. Our results show that they are effective in altering the latent representations of popular networks to improve their performance against adversarial attacks. 

The effectiveness of EITs in preventing adversarial attacks stems from their ability to roughly segregate the object from its background. In the process, EITs also cut through the some part of the object in their variably shaped blocks. This provides an additional reduction in features that could be redundant and overlap with other objects. This is analogous to creating a bottleneck in the feature space as done in popular deep networks for dimensionality reduction and feature selection~\cite{yu2011improved, yu2011improved, saxe2019information, pogodin2020kernelized, achille2018critical}. The movement of blocks from their original place also ensures breaking through some level of spatial biases giving rise to spurious correlations~\cite{wang2022deep, tivnan2022control}. Moving the pixels around leaves the background information in the input, contrary to learning with only silhouettes~\cite{baker2018deep}. The background helps with generalizability as depicted by suppressed activations in saliency maps. Studies have shown that generalization across multiple domains is hard to come by in a single network~\cite{tramer2019adversarial}, but EITs are able to achieve that without any computational overhead.

Our results also indicate the linearity of weight alteration and its effects on the performance against adversarial attacks. EITs are effective across all layers of the network and not only the readout. Training only the head of the network vs training the entire network shows the effectiveness of the weight changes in learning robust representations. In our experiments we found that changing the Imagenet-pretrained readout layers show only limited gains compared to pretrained networks, as shown by the diffused activations in saliency maps. The effect is more pronounced in networks with higher parameters (Figures~\ref{fig:saliency_efficientnetb0_fe_gn1}, \ref{fig:saliency_efficientnetb0_fe_gn3} and \ref{fig:saliency_efficientnetb0_fe_gn5}). The performance rises sharply when all weights of the Imagenet-pretrained network are retrained with EITs demonstrating the effectiveness of EITs in altering the latent representations at all layers. Training with EITs shows better perfromance compared to baselines in most cases, and follows very closely in others (could be easily ruled out as randomness). Hence, the performance gain is a function of training with EITs and not of validation with gaussian noise. 
% training only the head/readout vs full network
% EITs actually contribute to the performance as indicated by baseline v/s EIT trained

% our methods are effective, scalable and can be used with any off-the-shelf network
We show the effectiveness of EITs in learning robust latent representations against adversarial attacks. EITs are computationally inexpensive transformations and can be used in the data processing pipelines with any off-the-shelf network. Based on the neurophysiological findings, these transforms help with capturing the right set of features describing the object that would otherwise be prone to spurious correlations. The parameterization of EITs helps with capturing the right set of features from objects of different sizes. We found that the representations learnt with EITs are also generalizable across various different intensities of adversarial attacks. Our experiments show that EITs might be able to prevent black-box adversarial attacks with only limited finetuning and very little computational overhead. While the idea of describing an \text{``object"} might be hard, EITs show that it might be possible to learn the concept of \text{``objectness"} similar to how the human visual system processes it and can be used for object recognition in the wild.

% can combine with discussions

% what could be the reason for difference in performances with different EITs? -- explain in discussions.
% Also, do the experiments generalize to different gaussian noise levels?

\section{Limitations and Future Work}
% adding selective noise to only the salient parts of input (both from our pretrained networks and from the standard SOTA networks) and see how the pretrained network performs -- form of white box adversarial attacks.
% can the block sizes and prob of shuffle be learned as a paramter by a meta learning network?

Due to limited computational resources we were unable to complete a few experiments, including the segmentation transforms on EfficientNetB0. We were also able to validate with only gaussian noise corruption. In a future work, we would like to see how EITs would impact the latent representations of recurrent networks, and if EIT trained recurrent networks can be used for producing robust object segmentations. We would also like to see how does validating on one kind of corruption generalize to other kinds of corruptions and their intensities (for \eg validating with gaussian noise but testing on shot noise).

We would like to further extend our experiments in the white-box adversarial attacks space. We would like to add selective noise to only the salient parts of the input. These parts could be found by the activations in saliency maps of state-of-the-art networks based on the regions these networks attend to. We would also like to use our EIT trained/finetuned networks to see if altering the regions identified by activations in our saliency maps will lead to a performance degradation. This would help us probe the limits of the automatic feature selection done by EITs. 

An advantage of EITs is their ability to be parameterized. While the parameterization for EITs is fairly straightforward with only 3 parameters, the range in individual parameters can sometimes be hard to find, similar to the parameters in other kinds of input transformations. We would like to explore if these parameters can be learned and adjusted for different inputs using meta learning. 

% \section*{Acknowledgements}

%%%%%%%%% REFERENCES
{\small
\bibliographystyle{ieee_fullname}
\bibliography{egbib}
}

% %%%%%%%%%%%%%%%%%%%%%%%%%%%%%%%%%%%%%%%%%%%%%%
% %%%%%%% supplementary materials
% %%%%%%%%%%%%%%%%%%%%%%%%%%%%%%%%%%%%%%%%%%%%%%

% %%%%%%%%%% Merge with supplementary materials %%%%%%%%%%
% % \widetext
% \clearpage
% \setcounter{figure}{0}
% \setcounter{table}{0}
% \setcounter{section}{0}

% \makeatletter 
% \renewcommand{\thesection}{S\@arabic\c@section}
% \renewcommand{\thefigure}{S\@arabic\c@figure}
% \renewcommand{\thetable}{S\@arabic\c@table}
% \makeatother

% % \title{Supplementary Materials}
% % \noindent{\Large \textbf{The Challenge of Appearance-Free Object Tracking with Feedforward Neural Networks \\ \\ -- Supplementary Information --}}
% \twocolumn[  
%     \begin{@twocolumnfalse}
%         \begin{center}

%              \Large \textbf{Title \\-- Supplementary Information --}

%          \end{center}
%      \end{@twocolumnfalse}
% ]

\end{document}